\documentclass[a4paper, final]{article}
\usepackage{amsmath}
\usepackage{caption}
\usepackage{epigraph}
\DeclareCaptionType{capeq}[figure][List of equations]
\usepackage[dvips,pdftex]{graphicx}
\usepackage{multirow}
\usepackage{booktabs}
\usepackage{url}
\usepackage{hyperref}
\usepackage[numbers]{natbib}
\usepackage{authblk}
\title{Transforming Sensitive Documents into Quantitative Data: An AI-Based Preprocessing Toolchain for Structured and Privacy-Conscious Analysis}
\author{Anders Ledberg\thanks{anders.ledberg@su.se or anders.ledberg@gmail.com}}
\author{Anna Thalén}
\affil{Department of Public Health Sciences, Stockholm University SE-106 91 Stockholm, Sweden}
\date{\today}

\begin{document}
\maketitle

\begin{abstract}
Unstructured text from legal, medical, and administrative sources offers a rich but underutilized resource for research in public health and the social sciences. However, large-scale analysis is hampered by two key challenges: the presence of sensitive, personally identifiable information, and significant heterogeneity in structure and language. We present a modular toolchain that prepares such text data for embedding-based analysis, relying entirely on open-weight models that run on local hardware—requiring only a workstation-level GPU and supporting privacy-sensitive research.

The toolchain employs large language model (LLM) prompting to standardize, summarize, and, when needed, translate texts to English for greater comparability. Anonymization is achieved via LLM-based redaction, supplemented with named entity recognition and rule-based methods to minimize the risk of disclosure. We demonstrate the toolchain on a corpus of 10,842 Swedish court decisions under the Care of Abusers Act (LVM), comprising over 56,000 pages. Each document is processed into an anonymized, standardized summary and transformed into a document-level embedding. Validation—including manual review, automated scanning, and predictive evaluation—shows the toolchain effectively removes identifying information while retaining semantic content. As an illustrative application, we train a predictive model using embedding vectors derived from a small set of manually labeled summaries, demonstrating the toolchain’s capacity for semi-automated content analysis at scale.

By enabling structured, privacy-conscious analysis of sensitive documents, our toolchain opens new possibilities for large-scale research in domains where textual data was previously inaccessible due to privacy and heterogeneity constraints.
\end{abstract}

\section{Introduction}
Unstructured text from legal, medical, and administrative contexts offers a rich yet often underutilized source of information for research across domains such as public health, social science, and law \cite[e.g.,][]{Grimmer&Stewart2013}. Such texts often contain detailed accounts of individual circumstances, institutional reasoning, and administrative decisions—information that is typically absent from, or only partially reflected in, available structured data. For example, court records provide comprehensive descriptions of the actors and events involved in legal proceedings—information that can be valuable for characterizing judicial processes or predicting future outcomes, such as recidivism. Recent advances in text embeddings—dense vector-based representations of language that capture both semantic and structural properties—have enabled large-scale computational analysis of such texts \cite{Mikolov2013, devlin2019BERT, Reimers&Gurevych2019, Hou&Huang2025}. The goal of generating text embeddings is to transform unstructured textual data into numerical representations that encode salient linguistic features. Typically, each document in a corpus is represented as an $N$-dimensional vector, where $N$ may be in the order of $10^3$. This representation enables the use of a wide range of analytical tools developed for numerical data, such as similarity search \cite[e.g.,][]{karpukhin2020dense} (e.g., “find the $k$ most similar texts to a given one”) and both supervised and unsupervised classification methods \cite[e.g., ][]{Grootendorst2022BERTopic}.

Crucially, embedding vectors must accurately capture the salient aspects of the texts relevant to the analytic task at hand. In practice, source texts may be heterogeneous in both form and content. When such heterogeneity is unrelated to the information of interest, it is often beneficial to standardize the texts prior to embedding. Additionally, embedding models typically perform best on languages they were extensively trained on; even multilingual models may yield better representations for languages that are more prevalent in their training data. To address these issues, we propose using large language models (LLMs) to preprocess the textual data before embedding. LLMs are proficient at summarizing texts and translating them into languages that are well-supported by available embedding models. While these preprocessing steps enhance the analytical utility and comparability of the text data, a separate and equally important consideration is the protection of individual privacy within the documents.

Privacy protection is especially critical when working with legal and medical records, which often contain personally identifiable and sensitive information, such as details of individuals’ criminal or medical histories. Personal and sensitive data are strictly regulated under laws such as as the EU’s  General Data Protection Regulation (GDPR), requiring researchers to implement appropriate safeguards. This commonly entails pseudonymization or anonymization of the texts. Such measures are crucial even in the context of text embeddings, since information from the original texts may be partially recoverable from embedding vectors \cite{Song&Raghunathan2020, Li-etal2023, Morris-etal2023}.

Here we present a toolchain that transforms sensitive, unstructured text documents into standardized, anonymized, and embeddable summaries. The toolchain is designed to run entirely on local hardware using open-weight models, requiring only a single workstation-level GPU—making it suitable for privacy-sensitive research environments. To demonstrate its utility, we apply the tool\-chain to a corpus of 10,842 legal documents containing decisions from Sweden’s administrative courts, comprising over 56,000 pages of text. The primary output is a set of document-level embedding vectors suitable for statistical modeling, classification, and retrieval. Although this work focuses on legal documents, the approach is general and may be of interest to applied researchers working with sensitive texts in other domains, such as medical records \cite{lehman2021doesbert}.

As a practical illustration, we use a small set of manually labeled summaries to train a predictive model on the corresponding embedding vectors for identifying suicide-related content, demonstrating how embeddings can be used for scalable, semi-automated content analysis.

\section{Corpus}

The dataset comprises all court decisions (\textit{domar}) issued under the Swedish Care of Abusers Act (Lag (1988:870) om vård av missbrukare i vissa fall, hereafter LVM) by the lower administrative courts (\textit{förvaltningsrätter}) in Sweden between 2012-07-01 and 2024-06-30.  Ethical approval to work with these data was obtained from the Swedish Ethical Review Authority (Etikprövningsmyndigheten, application numbers 2023-07807, and 2024-05484-02). Under the LVM statute, a municipal social board is required to apply to the court for compulsory care if certain legal criteria are fulfilled—specifically, if an individual’s substance use poses a serious threat to their own health or safety, or to the safety of others. These legal documents are written in Swedish and originate from all 12 regional administrative courts in Sweden, covering diverse geographical and institutional contexts. 

In accordance with the Swedish principle of public access to official records (\textit{offentlighetsprincipen}), court decisions are, in principle, public documents and can be requested by any individual. These documents contain names and personal identity numbers, and often also sensitive personal information about the person ``on trial'' and potentially about other persons (e.g., close relatives). Anonymization of the documents before embedding is therefore essential. 

The relevant court decisions were obtained directly from the administrative courts in PDF format and subsequently converted to plain text using the \texttt{pdftotext} utility available in Linux systems, although any other pdf-to-plain-text converter would likely work as well.  The final corpus includes 10,842 court decisions, corresponding to 56,597 pages of text. The documents vary considerably in length and detail, reflecting differences in writing style, local practices, and individual case complexity. 

\section{Toolchain Overview}

The preprocessing toolchain consists of three main stages. In the first stage, a large language model (LLM) is prompted to summarize, anonymize, and translate each document. In the second stage, the resulting summaries are further scrutinized, and any remaining identifying entities are removed. In the third stage, each processed summary is transformed into a fixed-length vector using a text embedding model. This produces a set of document-level embedding vectors suitable for downstream machine learning or statistical analysis.

The toolchain is implemented in Python and is designed to run entirely on local hardware, using open-weight models.  All key steps are modularized to allow substitution of models, languages, or document types. This ensures transparency, reproducibility, and compatibility with privacy-sensitive data environments. A Python implementation is provided at \texttt{github.com/aledberg/preprocess}. However, we are not able to openly share the original court documents, but will do so to researcher's having the required ethical permissions.

\section{Toolchain Components}
\subsection{Stage 1: Prompted Summarization, Anonymization, and Translation}
The first stage of the toolchain uses a locally hosted instance of a large language model to generate English-language summaries that are both anonymized and standardized. The model used is Mistral Small 3\footnote{\texttt{mistral-small:24b-instruct-2501-q4\_K\_M}}, from Mistral AI, accessed via the Ollama framework (\url{https://ollama.com}). This model was selected after empirical testing as the best-performing model that could be run efficiently on the available hardware: a single Nvidia RTX A5000 GPU with 24GB of VRAM. The model is an instruction-tuned variant designed for summarization and general-purpose generation tasks, making it well-suited for prompt-based control without fine-tuning.

Each input document is passed to the model via the Ollama API, which is called from Python and runs the model locally. The prompt instructs the model to: (1) summarize the document according to a given structure, (2) remove or redact identifying information, and (3) translate the resulting summary into English. The following prompt was used: 

\vspace{0.5em}
\noindent
\begin{quote}
  The text provided below is a court decision on the Swedish Care of Abusers (Special Provisions) Act (LVM). Please summarize the case. Focus on the person being on trial but do not mention his or her name, and please pay attention to details. Replace the name of the person on trial with N.N. throughout, do not include name or personal number in your reply. Please answer in English using plain text (not markdown) and use the following format for your response:

  Administrative details: Date (of decision or trial); Location of the court; Type of trial; Outcome

  Personal details: Age  (can be obtained from the personal identification number (YYMMDD-NNNN)); Sex

  Substance use: Types of substances (including alcohol) used; History of use; Multiple substance use; Overdoses

  LVM history: Previous LVM episodes; Current status, is the person detained according to 13 § LVM (omedlebart omhändertagande)?.

  Physical health issues: (chronic medical conditions; acute health episodes requiring health care attention)

  Mental health issues: (previous care for psychiatric problems; previous compulsory care for psychiatric problems (LPT); suicide attempts and/or ideation

  Social conditions: (housing status; employment status; family situation)

  Arguments by the social board for why LVM-care is needed:

  Arguments from the person on trial (if present):

  [The Swedish court document goes here]
\end{quote}

\noindent
The prompt was developed iteratively by experimenting with different instructions and qualitatively inspecting the model’s outputs. Once a version produced summaries that were reasonably accurate, well-formed, and free of obvious identifying information, it was adopted for full-scale processing. More extensive prompt optimization was not performed due to processing time constraints: generating a summary for a single document typically took around 15 seconds.

This approach was chosen to maximize generalizability and ease of use: by relying on an instruction-tuned, open-weight model and prompt engineering rather than fine-tuning, the method remains lightweight and adaptable to other document types or languages. However, it is possible that fine-tuning a model for this task would lead to better performance. 

\vspace{0.5em}
\noindent
\textit{Note on scalability:} As both models and consumer-grade hardware continue to improve, the quality, speed, and accuracy of this step are expected to increase substantially. Future instruction-tuned models may improve linguistic consistency, and offer greater control over output formatting. Moreover, improvements in model capability and prompt design may eventually eliminate the need for a separate postprocessing step, as anonymization could be performed reliably during the generation process itself.

\subsection{Stage 2: NER-Based Postprocessing and Rule-Based Redaction}
The second stage of the toolchain applies additional filtering to ensure that no personally identifiable information remains in the LLM-generated English-language summaries. Named entity recognition (NER) is performed using the Stanza NLP library \cite{qi2020stanza}, configured to detect entities of type \texttt{PERSON} using the Swedish language model. The goal is to identify any names that may have been retained or introduced during the summarization step.  To avoid over-filtering, the detected names were manually reviewed. Some entities labeled as \texttt{PERSON} by the NER model corresponded to legitimate, non-identifying terms, including the names of institutions (e.g., \textit{Capio Maria}) or medical conditions (e.g., \textit{Wernicke-Korsakoff syndrome}). Such terms were retained in the summaries to preserve semantic integrity. In contrast, personal names not linked to identifiable institutions or diagnoses were removed or replaced with a neutral placeholder (\texttt{[NAME]}).

In addition to name-based filtering, regular expressions were used to remove structured patterns corresponding to other common identifiers. These included:
\begin{itemize}
    \item Street addresses, typically of the Swedish form \textit{[Name]gatan NN}, e.g., \textit{Storgatan 12B}, \textit{Karlavägen 3}
    \item Swedish personal identity numbers, typically of the form \texttt{YYYYMMDD-NNNN}
    \item Full birth dates. In such cases, the day of birth was removed, but the year and month were retained where available to preserve potentially relevant demographic information.
\end{itemize}

This combination of NER-based and rule-based redaction offers a layered safeguard, helping to ensure that residual identifiers are removed while maintaining as much non-sensitive content as possible for downstream analysis.

\subsection{Stage 3: Embedding Generation}
The final step in the toolchain involves transforming the anonymized English summaries into fixed-length vector representations using a large language model-based text embedding model. For this purpose, we used the \texttt{gte-Qwen2-1.5B-instruct} model \cite{li2023towards}, selected based on its strong performance on the Massive Text Embedding Benchmark (MTEB) leaderboard\footnote{\url{https://huggingface.co/spaces/mteb/leaderboard}}. This was the highest-ranked model on the benchmark that could be run efficiently on local hardware. 

The model was accessed via the Python \texttt{sentence-transformers} library. The resulting embeddings are 1536-dimensional vectors that capture the semantic content of the summaries. Embedding generation took approximately 0.15 seconds per document. These vectors constitute the primary output of the toolchain and are suitable for downstream tasks such as classification, clustering, and retrieval.
\section{Validation and Evaluation}
To assess the effectiveness of the preprocessing toolchain, we conducted two types of validation. First, we evaluated whether the anonymization procedures successfully removed personally identifiable information from the documents. Second, we examined whether the summarization and translation process preserved relevant semantic content from the original (Swedish) texts.

\subsection{Anonymization Validation}
To assess the effectiveness of the anonymization process, a two-part validation strategy was employed: manual review of a sample and automated scanning of the full dataset.

First, a random sample comprising five percent of the anonymized summaries (542 cases) was manually reviewed by the authors. Each summary was read in full to confirm that no directly identifying information remained. This included the full names of individuals, addresses, personal identity numbers, and any rare descriptors that could potentially enable re-identification.

Second, a full-corpus scan was conducted using regular expressions and fuzzy matching to detect residual personal information. In particular, each document was searched for the full name and personal identity number of the individual on trial, as recorded in the metadata accompanying the original court decisions. This automated check was designed to identify cases where personally identifying details may have inadvertently been preserved by the large language model and missed by the NER system and rule-based filters.

Together, the manual and automated checks confirmed that the combination of LLM prompting, NER-based filtering, and regular-expression-based redaction provided a high level of anonymization across the corpus. No identifying entities were found in any of the 542 reviewed summaries. The full-corpus search did not find any names. One case was identified in which a personal identity number had been retained; it was consequently removed.  

\subsection{Information Retention Validation}
Beyond anonymization, the LLM also translated the documents into English and summarized them—steps that impact the information retained. Translation was performed to optimize the quality of the embeddings, since the embedding model had not been trained on Swedish. The summarization serves two main purposes: to focus on comparable information across cases and to standardize the texts in both form and length.

As a coarse indicator of text reduction, we computed the mean and coefficient of variation (CV) for both character and word counts in the original documents and in the LLM-generated summaries. On average, the summaries were significantly shorter than the originals, reducing the overall length by approximately 75\% in terms of characters and 70\% in terms of words. Summaries were also less variable in length as indicated by a lower CV. For characters, the CV was 0.32  for original texts and 0.16 for summaries. For word counts, the CVs were 0.32 for the original texts and 0.17 for the summaries. 

\subsubsection{Information Retention}
To evaluate whether the anonymization, translation, and summarization process preserved information relevant for downstream tasks, we conducted a series of predictive modeling experiments using the document embeddings. Both the original Swedish documents and the anonymized English summaries were embedded using the same model—\texttt{gte-Qwen2-1.5B-instruct}—to ensure comparability.

Each document was labeled based on the presence of specific keywords or metadata fields, and a set of classification tasks was constructed to test whether these labels could be predicted from embeddings. The tasks were: (1) predicting the location of the court (12 possible classes), (2) predicting the year of the trial (13 possible classes), (3) detecting references to specific narcotic substances: amphetamine (in 21\% of documents), heroin (in 10\% of documents), and (4) identifying mentions of suicide attempts by keyword search (in 8.5\% of documents). All keyword-based labels were derived from the original Swedish texts.

For each task, L2-regularized logistic regression model were trained using the \texttt{scikit-learn} implementation. First, the regularization parameter \( C \) was optimized via grid search with five-fold cross-validation on the full data from the two embedding-types separately.  Second, model evaluation was performed, following a repeated train/test protocol: in each of 10 iterations, a random 75\% sample of the data was used for training and the remaining 25\% for testing. Performance metrics (e.g., accuracy and AUC) were averaged across the repetitions. The aim was to compare the performances between embeddings of the original texts and those of the anonymized summaries.

For the multiclass tasks (court and year), model performance was evaluated using overall classification accuracy and macro-averaged AUC. For the binary keyword detection tasks--where the positive classes were relatively rare--we used area under the precision-recall curve (PR-AUC), which is more informative than ROC-AUC when the outcome is imbalanced. PR-AUC focuses on the tradeoff between precision and recall in the positive class, providing a better sense of classifier quality in the context of rare-event detection. Table~\ref{table:1} shows the results of these tasks.

\begin{table}[h]
\centering
\caption{Predictive performance comparison between embeddings of the original Swedish texts and the anonymized English summaries across five classification tasks. Each metric is averaged over 10 repeated 75/25 train/test splits, with 95\% confidence intervals shown in parentheses.}
\label{table:1}
\begin{tabular}{lccc}
\toprule
\textbf{Task} & \textbf{Metric} & \textbf{Original Text} & \textbf{Summarized Text} \\
\midrule
Court location & Accuracy (\%)     & 99.5 (99.4, 99.6) & 97.7 (97.5, 97.9) \\
               & AUC (macro)       & 1.0 (1.0, 1.0) & 0.999 (0.998, 0.999) \\
Year of trial  & Accuracy (\%)     & 83.6 (83.3, 83.9) & 96.9 (96.6, 97.1) \\
               & AUC (macro)       & 0.986 (0.985, 0.986) & 0.996 (0.996, 0.996) \\
Amphetamine    & PR-AUC       & 0.88 (0.88, 0.89) & 0.97 (0.97, 0.97) \\
Heroin         & PR-AUC            & 0.84 (0.83, 0.86) & 0.93 (0.92, 0.94) \\
Suicide attempt & PR-AUC           & 0.48 (0.47, 0.50) & 0.60 (0.58, 0.62) \\
\bottomrule
\end{tabular}
\end{table}

The results showed that models trained on embeddings of the anonymized summaries outperformed those trained on embeddings of the original texts on all tasks except court location. Further, the excellent performance on predicting location and year of trial demonstrates that the embeddings indeed capture structural information exceedingly well. Together, these tests suggest that the summarization, translation, and anonymization process not only preserved core semantic content, but may have improved signal quality by reducing noise, irrelevant detail, or language variation. Note, however, that these results are most likely model dependent -- future models, better at embedding Swedish texts directly, could make the summarization and translation steps obsolete.  

\section{Application: Scalable Prediction of Suicide-Related Content from Limited Manual Annotations}

The anonymized summaries and embedding vectors produced by the toolchain are intended for downstream use in research, classification, and policy analysis. To illustrate one such application, we developed a predictive model for identifying cases that include references to suicide attempts or suicidal ideation.

\subsection{Background and Motivation}

Clients committed to care according to LVM face a high risk of death after discharge \cite{Ledberg&Reitan2022}. Suicide risk is a critical concern in this context, and early identification of at-risk individuals is of both clinical and public health importance. However, manually reviewing thousands of legal documents for references to suicide-related content is impractical. We therefore explored whether a model trained on a small, manually coded subset of summaries could generalize to the full corpus.

\subsection{Manual Annotation Protocol}

A random sample of 550 anonymized English summaries was read and annotated by the authors. Each summary was labeled as either containing (1) or not containing (0) indications of prior suicide attempts or suicidal ideation. Unclear cases were discussed and resolved to produce a final consensus label for each case.

\subsection{Predictive Modeling Approach}

The labeled dataset was used to train a logistic regression model with L2 regularization, using the 1536-dimensional document embeddings as input features. The model was evaluated using 5-fold cross-validation, yielding a PR-AUC score of 0.8. The optimal value of the regularization parameter (C) was around 10. Note that with 550 observations and 1536 predictors, it is essential to use regularization for useful predictive power. 

\subsection{Corpus-Wide Prediction}

After training, the model was applied to the remaining 10,292 summaries in the corpus. For each document, the model returned a predicted probability between 0 and 1 indicating the likelihood that the summary contained references to suicide attempts or ideation. These probabilities provide a scalable proxy for manual annotation and can be used in downstream statistical or epidemiological analysis, either as a binary indicator (after thresholding) or as a continuous measure of suicide-related content.

\subsection{Post-Hoc Validation of Predictions}

To assess the quality of predictions beyond the initial training set, we conducted a follow-up validation on 500 additional summaries not used in model development. We randomly selected 250 cases that the model had assigned a probability less than 0.5, and 250 cases with a predicted probability more than 0.5. Each of these summaries was read by one of the authors (AT) and manually assessed for suicide-related content using the same criteria as in the initial labeling phase.

\begin{figure}[htbp]
  \centering
  \includegraphics[width=0.7\textwidth]{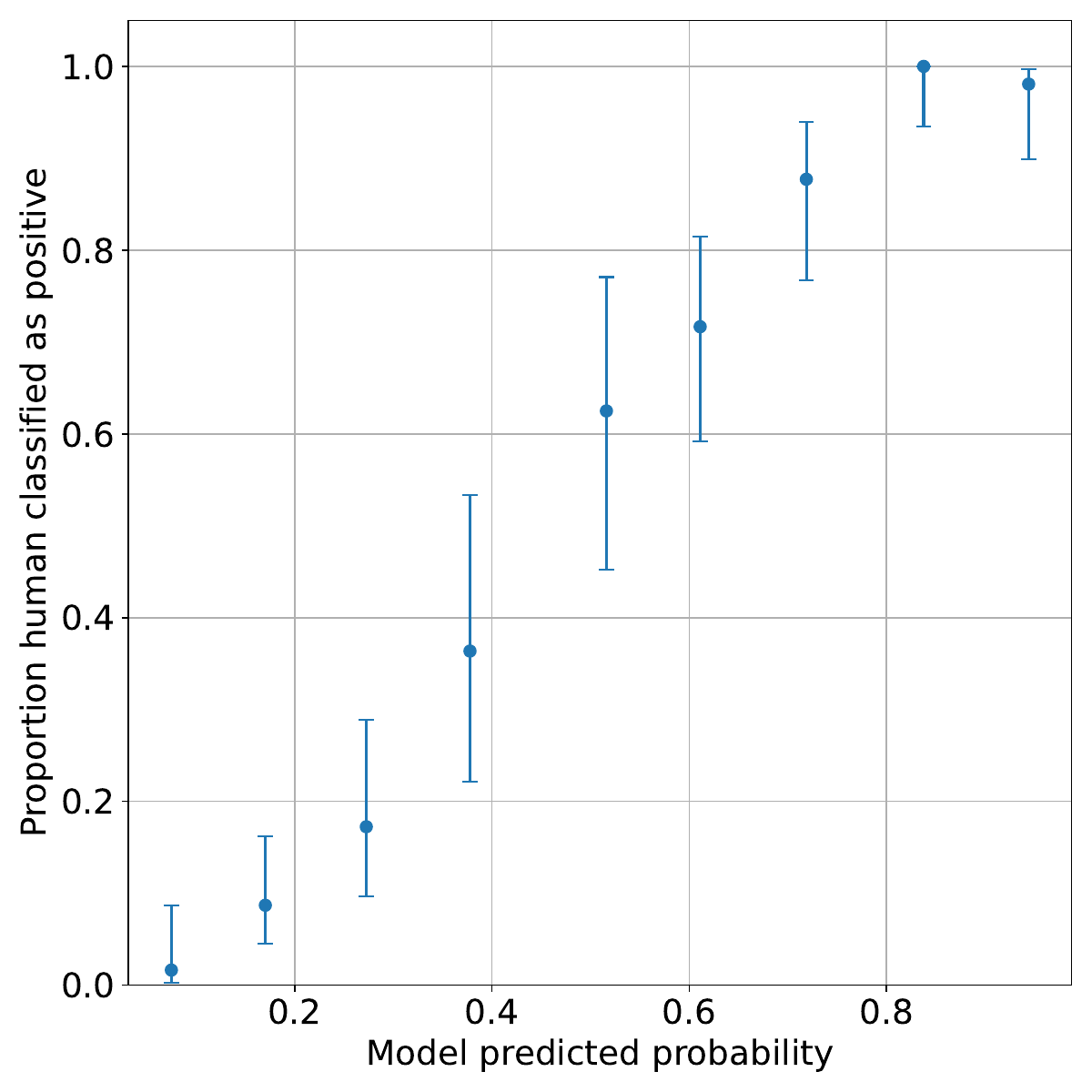}
  \caption{Comparison of predictions from the ML-model with classifications from a human observer. Five hundred summaries were read and classified by one of the authors (AT). The predictions from the logistic regression model were divided into nine bins, and the plot shows the average human classification for each bin with error bars representing 95 percent confidence intervals (computed by the Wilson score intervals formula). }
  \label{fig:1}
\end{figure}

The consistency between model predictions and human coding across these samples are shown in Fig~\ref{fig:1}. The figure shows a strong monotonic relationship between predicted probability and observed rate of suicide-related cases, indicating that the model generalizes well to unseen data. This post-hoc validation provides further evidence that the embeddings capture semantically meaningful features related to suicide-risk, even in the absence of explicit keywords.

\subsection{Interpretation and Implications}

This application demonstrates how the anonymized summaries and embeddings produced by the toolchain can be used to extend human judgment across large datasets. Compared to manual review, the approach is more scalable and may even improve generalization by capturing semantic patterns that are not explicitly recognized during annotation. It also highlights the value of document-level embeddings for behavioral and mental health research using legal text data. In future work, we aim to investigate whether these predicted probabilities are associated with actual suicide attempts occurring after discharge from the LVM-care, thereby assessing the model’s value for prospective risk stratification and public health surveillance.

\section{Concluding discussion}
This study presents a modular, scalable, and privacy-conscious toolchain for transforming sensitive texts into structured embedding representations suitable for downstream analysis. By integrating large language model (LLM)-based prompting, named entity recognition (NER), and high-dimensional embeddings, the toolchain offers a practical solution to the dual challenges of data anonymization and standardization—two essential prerequisites for embedding-based analysis of unstructured text.

A central strength of the proposed approach lies in its modularity. Each component—LLM, NER engine, and embedding model—can be independently replaced or upgraded without altering the rest of the pipeline. This design makes the toolchain adaptable to evolving needs across domains, languages, and use cases. For instance, newer LLMs can be substituted as they become available, enabling higher-quality summarization or more nuanced anonymization. Likewise, domain-specific NER models or multilingual embedding models can be integrated to better support corpora beyond the current focus on Swedish legal texts. This plug-and-play architecture facilitates ongoing refinement and broader applicability.

The reliance on prompt-based summarization and anonymization --rather than model fine-tuning-- is a design choice that promotes accessibility. Prompting enables researchers with limited machine learning expertise to harness powerful LLMs through natural language instructions, significantly lowering the technical barrier to entry. While fine-tuned models might improve task-specific performance, they are not required for effective use of the toolchain. On the contrary, users with access to task-specific models can readily integrate them into the existing framework, highlighting the toolchain’s openness rather than constraint.

The validation results show that the toolchain achieves a high degree of anonymization while preserving semantic content relevant to analytical tasks. In fact, embeddings of the standardized English summaries often performed better than embeddings of the original texts in predictive tasks. This suggests that controlled summarization and translation can enhance signal quality by removing irrelevant variation and linguistic noise. Such preprocessing may be particularly beneficial when working with heterogeneous corpora or languages underrepresented in LLM training data.

The suicide-related content prediction case study further demonstrates the toolchain’s utility in real-world applications. With relatively minimal manual labeling, a model trained on document embeddings generalized well to unseen cases, supporting scalable content analysis of behavioral and mental health indicators in legal texts. This kind of semi-automated annotation can enable new forms of public health surveillance, policy evaluation, and risk modeling based on unstructured administrative data.

Naturally, results from this demonstration cannot be assumed to transfer identically to other text types, legal systems, or languages. However, the architecture is explicitly designed to be generalizable. As long as the core steps—summarization, anonymization, and embedding—can be adapted to the language and structure of the source documents, the toolchain remains applicable. In multilingual contexts, translation may still be required, but can be adjusted to target a language that is well-supported by embedding models.

In conclusion, the toolchain provides a transparent, flexible, and effective means of preparing sensitive unstructured texts for embedding-based analysis. Its modularity ensures adaptability, its reliance on prompting enhances accessibility, and its validation suggests strong performance even in complex domains. Future work may explore integration with differential privacy methods, adaptation to other languages or document types, and expanded downstream use cases.

\section{Acknowledgments}
This research was funded by Statens institutionsstyrelse (SiS), the Swedish National Board of Institutional Care, grant number 2.6.1-1656-2022.
\section{Conflict of interest statement}
On behalf of all authors, the corresponding author states that there is no conflict of interest.
\section{Data availability statement}
The court documents analyzed in this study contain sensitive and personally identifiable information and therefore cannot be shared openly. Access to the data may be granted to qualified researchers with appropriate ethical approval and permissions, in accordance with Swedish regulations and institutional guidelines. Interested researchers are encouraged to contact the corresponding author to discuss data access arrangements.

\end{document}